# A MODIFICATION OF THE CONJUGATE DIRECTION METHOD FOR MOTION ESTIMATION.

Marcos Faúndez Zanuy, Francesc Tarrés Ruiz.
Signal Processing Department
EUETT La Salle.Enginyeria de Telecomunicació.
Ramon Llull University. Pg. de la Bonanova 8.
08022 BARCELONA (SPAIN)

## ABSTRACT

A comparative study of different block matching alternatives for motion estimation is presented. The study is focused on computational burden and objective measures on the accuracy of prediction.

Together with existing algorithms several new variations have been tested. An interesting modification of the conjugate direction method previously related in literature is reported. This new algorithm shows a good trade-off between computational complexity and accuracy of motion vector estimation.

Computational complexity is evaluated using a sequence of artificial images designed to incorporate a great variety of motion vectors. The performance of block matching methods has been measured in terms of the entropy in the error signal between the motion compensated and the original frames.

## I. INTRODUCTION

There are two main alternatives to compress an image. The first one, usually called intraframe approach, pretends to remove the spatial redundancy of an image without destroying important information. These methods are suitable for fixed image applications such as multimedia, image data base, photo-coding, etc. Nevertheless, in applications which use a sequence of image data such as TV scenes, Video conference etc, time redundancy can be exploited to increase the compression ratio since consecutive frames are usually highly correlated. This second group of methods is called interframe approach, and pretends to remove temporal redundancy. There are two main methods for this second approach: differential coding, and differential motion-compensated coding. In the second case, motion estimation of subimages must be computed.

This paper is focused on the study of motion estimation procedures for the latter class of coding.

Obviously, hybrid methods exploiting time and spatial redundancy achieve higher compression ratios at the expense of computational burden [8].

Differential coding consists on making the difference between consecutive images. Theoretically, one more bit is needed for coding the difference, but in the case of highly correlated images, where the first image is a good prediction of the second, the dynamic range of the image is reduced, and it is possible to encode the image with less bits.

The differential coding with motion compensation is a sophistication of the differential coding.

## I.1. DIFFERENTIAL CODING WITH MOTION COMPENSATION

Although there exist many alternatives for differential coding with motion compensation, the main steps of this coding scheme can be resumed:

1-. To split the image into subimages (called blocks), not overlapped, with a typical size of 8x8 or 16x16 pixels
2-. Estimate the motion experimented by every block from one frame to the next one. It is represented with a two dimensional vector.
3-. To construct the prediction moving the blocks as indicated by motion estimation described in step 2. There are three cases:
a) pixels with two or more estimated values: the value is the average of all of them.
b) pixels with only one value.
c) pixels without estimated values: in this case, they remain with the value of the previous image.
4-. To make the difference between the original image and the prediction.
5-. The result of the step 3 is the prediction error, which can be quantificated for a high compression factor.
6-. Save or transmit the motion estimation information plus the prediction error.

The most critical step is the second one, since it determines the quality of prediction and therefore the compression ratio.There are two basic classes of methods for motion estimation: Block matching algorithms, and recursive methods [5] [6] [2]. The latest try to minimize recursively the squared displaced frame difference using an algorithm like steepest descent or LMS. There are more possibilities, using a rational equation instead a polinomical equation [2].

In this paper only block matching methods will be considered.

## I.2. BLOCK MATCHING

When the size of the blocks (NxN), and the maximum displacement allowed (dm) are fixed, a search area can be defined. Block matching consists on finding the position of the block ,in the search area, that minimizes an error criterion. The procedure that consists on evaluating all the possible situations is known as Full Search. It requires the computation of $(2dm+N)^2$ positions. A considerable reduction is achieved with an intelligent search. This is realized with fast block matching algorithms, that reduce the search area progressively. [2],[3],[4],[7].

## I.3. PARAMETERS SELECTION

For a practical implementation of a block matching algorithm, a number of parameters must be selected. These parameters can be summarized as:

* Block size:

The block size determines a trade-off between estimation accuracy and number of displacement vectors. That means that as block size decreases, smaller subimages are compared and precise estimations can be obtained. Nevertheless, small block sizes imply a great number of estimations that increase computational complexity.

Typical values are 8x8 and 16x16. A previously study [1] showed that for 515x512 pixels images, a 16x16 block size achieves quite good performance. Higher compression ratios can be obtained with 8x8 pixels but at the expense of considerable computational effort. A size of 16x16 has been selected in this study.

* Error criterion:

Usual error criteria are mean square error (MSE) or mean absolute frame difference (MAD). If the former is implemented with tables to avoid the multiplication, the computational time is nearly the same. In this study MAD has been chosen as the error criterion, since differential coding is based on linear pixel differences and not on quadratic differences.

* Evaluation of the prediction quality:

Typical measures are entropy, variance of the error image, and signal to noise ratio.

The variance may be a non realistic measure. For instance, an image with only two contrasted levels in a set of 255 possible levels, has a great variance, and it is very simple to predict (only one bit is needed). The SNR has the same problem because it depends of the variance. For this reason, we choose the entropy.

## II. MODIFIED CONJUGATE DIRECTION ALGORITHM

As a result of an exhaustive study of the 2D logarithmic, three step, conjugate direction, and orthogonal search algorithms, some new variations of block matching algorithms were proposed. Most of them were a trivial simplifications of the classical algorithms, that concluded with no relevant results. Nevertheless, a modified version of the conjugate direction algorithm (CDA) was specially interesting for his properties.

CDA consists on estimating the motion on the X axis, calculating the error on the considered point and the left and right points. The point with lowest error is taken as the result of the present step. The process is repeated until the minimum is the central point examined or it is in the limits of the search area. Then, the process is repeated on the Y axis, examining the up and down points. After this process we have to possibilities:

1. The estimation is finished. Then, the algorithm is named One_aT_a_Step (OTS).
2. The algorithm tries to adjust again in the X axis, repeating the process until a minimum is found in the X and Y axes. Then, it is the Conjugate direction Algorithm (CDA).

Really, OTS is a simplified version of the CDA. Some studies (i.e. [1],[9]) show that there are not important differences between them. In our evaluation we chose the OTS because it is simpler and faster. Nevertheless we name our algorithm modified conjugate, because the original algorithm is the CDA.

Our comparative study of the 2-D logarithmic, three step, conjugate direction and orthogonal search algorithms revealed that the conjugate direction method had the better performance and the lowest computational complexity for estimating small displacements in real and artificial images. However, the method showed a poor behaviour (measured in entropy) when the displacement to be estimated is large. It was concluded that the different behaviour of the methods was mainly due to the spatial distance between the points examined in the same step. If the points are near, then the decision criterion is not clear since the errors are similar. However if the points are quite separated then the decision is easier because the errors are more contrasted. Besides, the conjugate direction method has a very simple structure decision. These factors let think on a modification to improve the performance for estimating large displacements, without incrementing the computational cost.

The way to speed up the conjugate direction method and to improve the entropy for long displacements, is incrementing the separation between examined points in the same step. An increment of two pixels is suitable for blocks of 16x16 pixels and a maximum displacement of six. When the minimum is found in the central point of the three examined points, then the increment is reduced to one pixel.

In figure 1 are represented the successive approaches for estimating the vector (2,6).

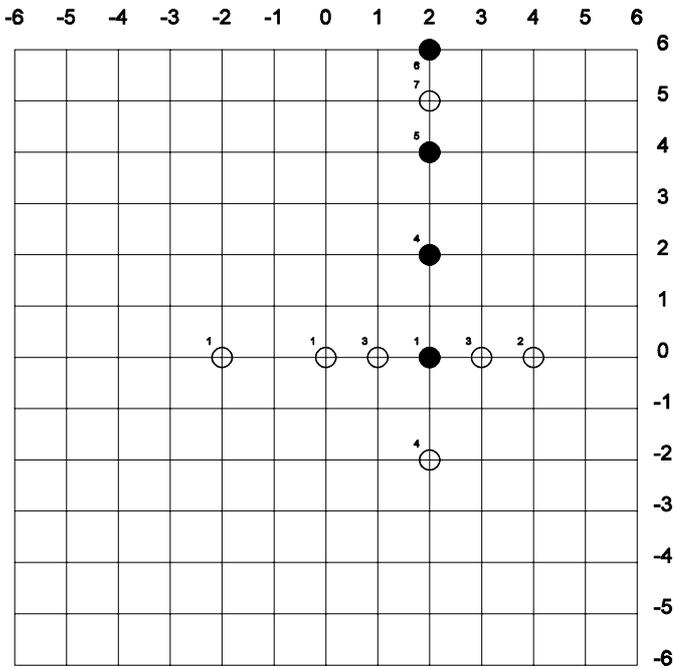

*Figure 1. To evaluate the vector (2,6), the method found the approaches (2,0), (2,2) and (2,4).*

The numbers denote the examined points in each step. Shadowed circles are the result of the step.

It is interesting to observe that in the first step, three points are calculated, but in the next steps only one is needed, because the others were calculated in the precedent step. (When the algorithm changes to Y axis, then two points are computed). This fact, joined with the simple structure decision is the reason of the speed of this algorithm.

There are two independent variations to the proposed method:
1. When the minimum is found in the X axis, the step is not reduced, and the algorithm searches for a minimum in the Y axis. Then, the step is reduced to one, and the final adjust is done in the X-Y axes. With this modification, the evaluation time is nearly the same, and the accuracy improves, since the final decision in the X axes is taken in the final steps, where it is less critical.
2. When the minimum is found in the central point, it is not necessary to examine the two neighbour points. The algorithm only needs the evaluation of the point between the central point, and the point separated two pixels, with lowest error: if the central point has coordinates (x,y), and the minimum error, there are two possibilities:
* $MAD(x+2,y) < MAD(x-2,y)$: then it is necessary the calculation of the point (x+1,y), and to choose the minimum between (x,y), (x+1,y)
* Otherwise, to compute (x-1,y) and choose the minimum between (x,y), (x-1,y).

In the Y axis the process is the same.

With this simplification, the number of evaluated points is the lowest of the block matching algorithms proposed so far. (11 points. Two less than the OSA). With this modification, the entropy is worse. In both cases, the method can detect all possible displacement vectors.

Obviously, both variations can be simultaneously incorporated in the algorithm.

## III. SIMULATIONS AND RESULTS
Our study about the different block matching algorithms for motion estimation includes the following features:
* Evaluation of the algorithm speed.
* Entropy.
* Graphic representation of the prediction error.

### III.1 COMPUTATIONAL TIME
So far, the way of evaluating the differences in computational complexity was the number of points required to find the displacement vector field in the worst case situation. This method does not allow to make comparisons with other alternatives such as recursive methods, image-flow approach, etc, and it is not a realistic measure, since real images are composed of complex movements and only a small fraction of them corresponds to the worst case situation. Theoretically, the better way to evaluate this, is by means of statistics. This is quite complex and not standard.

The number of evaluated points in the worst case situation and for a displacement vector (2,6) are summarized in table I.

| METHOD | POINTS | | STEPS | |
| --- | --- | --- | --- | --- |
| | A | B | A | B |
| FULL SEARCH | 169 | 169 | 1 | 1 |
| LOG. 2D | 18 | 21 | 5 | 7 |
| THREE STEP | 25 | 25 | 3 | 3 |
| CDA | 12 | 15 | 9 | 12 |
| ORTHOGONAL | 13 | 13 | 6 | 6 |
| MOD. CONJ. | 11 | 13 | 7 | 8 |

table 1.
A) Vector (2,6). B) Worst case situation.

Another way, is to form artificial images, with any type of predominant module or direction motion. This is due because it is possible to find an image that fits better to a predetermined method.

Artificial images have the following advantages:
* Exact control over the motion estimation quality and kind of motion.
* All the possible displacement vectors can be distributed around the image.

The proposed image, named star (512x512 pixels, 256 grey levels), is composed by black squares with a size of 16x16 pixels in a white background, forming an asterisk. There are four lines with the directions of the X, Y axes, and the two diagonals forming 45 degrees with the axes. Every line is formed by black squares spaced 16 pixels. The next image, called displaced star is formed with a progressive displacement of the squares from 0 to 7 pixels, giving an apparience of an expanding star, when they are visualized consecutively.

In table 2 are summarized the relative times required for each algorithm considered in our study. They are referred to the first block matching algorithm published, the 2D logarithmic search [3].

| METHOD | RELATIVE TIME (%) |
|---|---|
| 2D LOGARITHMIC | 100% |
| THREE STEP | 110.88% |
| CONJUGATE | 69% |
| ORTHOGONAL | 77.35% |
| FULL SEARCH | 498.27% |
| MOD. CONJUGATE | 60.2% |

Table 2.

It should be mentioned that simulation of any algorithms was stopped when in any iteration zero error was achieved. Other authors stop the research when the error is lower than a predetermined threshold.

A comparison between the tables shows a similar results, except for the case of the orthogonal search algorithm. Although it calculates less points, it takes more time than the CDA. This is due to the more complex decision structure of the OSA. For this reason, a fast algorithm must search a few points, and have a simple structure decision.

### III.2 ENTROPY

Two sets of real images were used to evaluate the entropy. The first one, is a non standard sequence of 18 images (256x256 pixels, 256 grey levels). The second one, is the Miss America sequence, and 24 images were used.

In this case, the entropy is very similar in all the images of the same sequences, so we only give the results between the first images. The results are represented in table 3.

| METHOD | ENTROPY | |
|---|---|---|
|  | SEQ1 | M.AMERICA |
| 2D LOGARITHMIC | 2.9092 | 3.5837 |
| THREE STEP | 2.8858 | 3.5438 |
| CONJUGATE | 2.8688 | 3.5972 |
| ORTHOGONAL | 2.8877 | 3.5602 |
| FULL SEARCH | 2.8708 | 3.5127 |
| MOD CONJUGATE | 2.8924 | 3.5702 |

table 3.

For the modified conjugate direction, the results do not include the variations proposed before. Further improvements are obtained if these modifications are considered. Nevertheless, in order to compare the performance of the modified conjugate method to other alternatives we preferred to neglect the influence of secondary factors that could also be incorporated into other alternatives.

The movement is greater in the second sequence, so the behaviour of the block matching algorithms is different for the two cases.

### III.3 CONCLUSIONS

To conclude we would like to remark the following features:
* There are few differences in entropy between methods, so is recommended to choose the fastest method.
* The results confirm the good behaviour of the CDA for small displacements, that grow worse for long displacements.

As concerns the graphic visualization of the prediction error the main conclusion is that there is a great difference between differential coding and differential coding with motion compensation, specially for the motion of undeformable objects: for instance, in an image with a women speaking, the error due to the body motion is just perfectly cancelled applying motion compensation

Another good characteristic of the proposed method, is the regularity in the computational time, independently of the type of motion (small, large, only in one axis or combined, etc). This is due to the near constant examined points (between 9 and 13). (Considering that the search does not finish when the error is lower than a predetermined threshold).